\pdfoutput=1
\documentclass[letterpaper]{article} 
\usepackage{aaai23}  
\usepackage{times}  
\usepackage{helvet}  
\usepackage{courier}  
\usepackage[hyphens]{url}  
\usepackage{graphicx} 
\usepackage{natbib}  
\usepackage{caption} 
\usepackage{algorithm}
\usepackage{algorithmic}

\usepackage{graphicx} 
\usepackage[justification=centering]{caption}
\usepackage{textcomp}
\usepackage[super]{nth}
\usepackage{xspace}

\usepackage{amssymb} 

\usepackage{newfloat}
\usepackage{listings}
\DeclareCaptionStyle{ruled}{labelfont=normalfont,labelsep=colon,strut=off} 
\floatstyle{ruled}
\newfloat{listing}{tb}{lst}{}
\floatname{listing}{Listing}
\title{Enabling High-Level Machine Reasoning with Cognitive Neuro-Symbolic Systems}
\author{
    Alessandro Oltramari{\rm}\\
}
\affiliations{
    \textsuperscript{\rm }Bosch Center for Artificial Intelligence\\

    2555 Smallman Street, Suite 302\\
    Pittsburgh, Pennsylvania 15222 USA\\
    alessandro.oltramari@us.bosch.com
}

\begin{document}

\maketitle

\begin{abstract}
High-level reasoning can be defined as the capability to generalize over knowledge acquired via experience, and to exhibit robust behavior in novel situations. Such form of reasoning is a basic skill in humans, who seamlessly use it in a broad spectrum of tasks, from language communication to decision making in complex situations. When it manifests itself in understanding and manipulating the everyday world of objects and their interactions, we talk about \textit{common sense} or \textit{commonsense reasoning}. State-of-the-art AI systems don’t possess such capability: for instance, Large Language Models  have recently become popular by demonstrating remarkable fluency in conversing with humans, but they still make trivial mistakes when probed for commonsense competence; on a different level, performance degradation outside training data prevents
self-driving vehicles to safely adapt to unseen scenarios, a serious and unsolved problem that limits the adoption of such technology.
In this paper we propose to enable high-level reasoning in AI systems by integrating cognitive architectures with external neuro-symbolic components. We illustrate a hybrid framework centered on \texttt{ACT-R}, and we discuss the role of generative models in recent and future applications.
\end{abstract}

\section{Introduction}
A large part of neuro-symbolic systems is based on transforming symbolic knowledge into sub-symbolic representations that are suitable for infusion in data-driven learning algorithms: Knowledge Graph Embedding (\textsc{kge}), among the others, is a prominent approach to reduce knowledge graph (\textsc{kg}) triples to latent vectors \cite{8047276}. Such transformation is instrumental to efficient computability of \textsc{kg} properties, as well as to application in a variety of downstream tasks: for instance, in \cite{wickramarachchi2023clue} the authors leverage \textsc{kge} methods to label unseen entities in autonomous driving datasets. Whether the \textsc{kge} process is realized by geometric, tensor or deep learning models, the purpose is to \textit{compress} \textsc{kg} structures into a low-dimensional space, where symbolic statements are replaced with dense, sub-symbolic expressions. Concatenation, non-linear mapping, attention-like mechanisms, gating mechanisms, are further methods to adapt knowledge structures to neural computations -- e.g., \cite{peters2017semi, strub2018visual, margatina2019attention}.\\
While knowledge-infusion can improve neural models, it is not sufficient to enable \textit{high-level reasoning}, which is typically required by complex tasks such as natural language understanding, activity recognition, decision making in complex scenarios: latent, sub-symbolic expressions can only augment training signals with features derived from explicit semantic content, but this infusion process does neither carry any information about the reasoning mechanisms needed to process the learned knowledge, nor instruct the neural models on how those should unfold. \\
But, what do we mean with \textit{high-level reasoning} and why is it important to endow artificial intelligent systems with such feature? \\

\section{Problem statement}

We can define \textit{high-level reasoning} as the capability to generalize over knowledge acquired via direct or mediated experience, and to exhibit robust behavior in novel situations. This definition is inspired by Kahneman's \textsc{system 2} mode of thought \cite{kahneman2011thinking}. When \textit{high-level reasoning} manifests itself in understanding and manipulating the everyday world of objects and their interactions, we talk about \textit{common sense} or \textit{commonsense reasoning}. State-of-the-art AI systems don't possess such capability: for instance, Large Language Models (\textsc{llm}s) have recently become popular by demonstrating remarkable fluency in conversing with humans, but they still make trivial mistakes when probed for commonsense competence (see next section); on a different level, one of the motivations why the promise of autonomous cars hasn't panned out yet concerns performance degradation outside training data, which prevents self-driving vehicles to safely adapt to unseen scenarios.\footnote{A main weakness of deep learning approaches, as stated in a recent article \cite{bengio2019meta}, is that `current methods seem weak when they are required to generalize beyond the training distribution, which is what is often needed in practice', such as in safely maneuvering a vehicle.} Humans, on the opposite, are very good at generalizing from a few examples, and at filling the gaps in experience with reasoning: for instance, when asked about what happens after a bottle of red wine is thrown against a concrete wall, even children can answer with the utmost certainty that the bottle will shatter and the wall will be wet and red-stained -- they can also easily infer that the impact between \textit{any} fragile material and \textit{any} hard surface \textit{typically} ends with the former being substantially altered, if not destroyed; analogously, student drivers only need limited training to learn how to safely maneuver a car, adapting their knowledge and skills to novel situations. Compared to current AI systems based on GPU accelerated computing, human reasoning capabilities are impressive, even more so when we factor in what Herbert Simon used to call `bounded rationality' \cite{simon1955behavioral}, i.e., the notion that human cognition operates with limited knowledge and is subject to time constraints -- a heritage of evolution \cite{santos2015evolutionary}.\\
As these arguments suggest, a cognitive stance toward designing AI systems \cite{lieto2021cognitive} seems to be key to enable high-level reasoning capabilities at the computational level: accordingly, we propose to complement cognitive architectures \cite{kotseruba202040, langley2009cognitive} with  neuro-symbolic methods. In this paper we illustrate the blueprints of a \textit{cognitive neuro-symbolic reasoning} system centered on the \texttt{ACT-R}\footnote{Abbreviation of `Adaptive Control of Thought, Rational'.} cognitive architecture \cite{anderson1996act}, whose hybrid (symbolic and sub-symbolic) mechanisms are well-suited for integration with neuro-symbolic algorithms and resources. Note that the proposed approach is applicable to any cognitive architecture whose properties are compatible with \texttt{ACT-R}, such as \texttt{SOAR} \cite{laird2019soar} and \texttt{SIGMA} \cite{rosenbloom2016sigma}: in fact, these three architectures have been grouped into the so-called `Standard Model of the Mind' \cite{laird2017standard}, an idea that has its roots in physics.\footnote{For a brief introduction to the \textit{Standard Model of Particle Physics}, see this resource from the U.S. Department of Energy:\\
\url{https://www.energy.gov/science/doe-explainsthe-standard-model-particle-physics}} Note that the Standard Model of the Mind doesn't prescribe how to implement cognitively-inspired AI systems; rather, it aims to play the role of a conceptual framework of reference for developing them.

\section{Motivations}
\begin{figure*}
    \centering
    \includegraphics[scale=0.88]{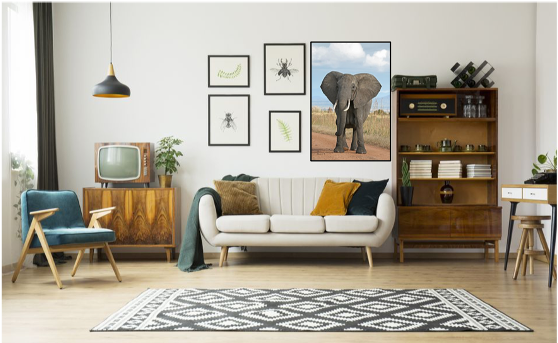}
    \caption{The \textit{Elephant in the Room}: the probability that a label assigned by an object detection system is correct increases when the context is factored in: in this example, the label `elephant' could plausibly denote a picture of the pachyderm, but not the pachyderm itself.}
    \label{fig:elephant}
\end{figure*}

Over the last decade, deep learning has yielded tremendous advancements in many AI fields, such as computer vision. For instance, neural models can achieve high accuracy in object detection when training and testing domains originate from the same data distribution. However, recent work shows that minimal/regional modifications implanted in the data at test time cause significant drop in accuracy \cite{eykholt2018robust, rosenfeld2018elephant}. The examples documented in \cite{rosenfeld2018elephant} are of particular interest, as they indicate how commonsense contextualization, by means of incorporating a priori structured knowledge into deep networks, can mitigate the effect of those perturbations, resulting in more robust performance \cite{marino2016more}. In general, a visual model suitably infused with knowledge extracted from semantic resources like \textsc{conceptnet} \cite{speer2017conceptnet} can strengthen the connections holding within instances of the same conceptual domain (e.g., \textit{couch}, \textit{television}, \textit{table}, \textit{lamp} are located in living rooms) and discard out-of-context interpretations (e.g., no real \textit{elephants} are located in living rooms, but photographs of elephant may be -- figure \ref{fig:elephant} depicts such case).  

When shifting to natural language, and to tasks like automated question answering, the key role played by knowledge-based contextualization for neural language models stands evident.\footnote{We use `language model', `neural language model' and `large language model' as interchangeable terms, as they commonly refer to the same neural architecture based on multi-headed self-attention mechanisms \cite{vaswani2017attention}; however, computational power significantly differs as function of the specific implementations (e.g., \textsc{bert} has 6 blocks with 12 heads, \textsc{gpt-3} has 24 blocks and 48 heads), and of the size of training datasets (\textsc{chat-gpt} has been trained on a massive corpus -- 570 GB -- of text data).} For instance, it has been demonstrated that using \textsc{kg} triples to disambiguate textual elements in a sentence, and embed the corresponding concepts and relations in neural language models \cite{devlin2018bert}, significantly improves performance \cite{ma2021knowledge}. In fact, despite of the impressive results that \textsc{llm}s are producing in Natural Language Processing \cite{ma-etal-2019-towards, bauer-bansal-2021-identify, shwartz-etal-2020-unsupervised}, basic reasoning capabilities are still largely missing. This is also the reason why it's not appropriate to use `Natural Language Understanding' to denote these tasks, because it would entail that robust and comprehensive reasoning capabilities are present \cite{mcshane2017natural}. Let's expand on this argument and consider a few representative examples. \\
In ProtoQA \cite{boratko2020protoqa}, \textsc{gpt-2} \cite{dale2021gpt}  fails to select options like `pumpkin', `cauliflower', `cabbage' as top candidates, for the question `one vegetable that is about as big as your head is?': instead, `broccoli', `cucumber', `beet', `carrot' are predicted. In this case, the different models learn some essential properties of vegetables from the training data, but do not seem to acquire the capability of comparing their size to that of other types of objects, revealing a substantial lack of \textit{analogical reasoning} \cite{ushio2021bert}. The same issues are observed when \textsc{chatgpt}, a recent popular version of \textsc{gpt-3} optimized for conversations, is considered: the main difference is that \textsc{chatgpt} is capable of generating plausible answers when the question is submitted literally, but often fails to do so when the verbal expression `about as big as' is paraphrased with alternative forms like `about the same size', `about the same shape', `comparable to', etc. This `hypersensitivity' to surface-level linguistic features -- an epiphenomenon of the model's incapability to generalize over textual variations of the same content -- seem to indicate that the model cannot perform the necessary (analogical) reasoning steps needed to correctly answer to the question. Along these lines, recent work \cite{ettinger2020bert} has shown that lack of complex inferences, role-based event prediction, and understanding the conceptual impact of negation, are some of the weaknesses diagnosed when \textsc{bert} \cite{devlin2018bert}, one of prominent open source language models, is applied to benchmark datasets. ProtoQA again provides good examples of these deficiencies: in general, neural models struggle to correctly interpret the scope of modifiers like `not' (\textit{reasoning under negation}), `often' and `seldom' (\textit{temporal reasoning}). Regarding the latter, in task 14 of bAbI \cite{weston2015towards}, a comprehensive benchmark challenge designed by Facebook Research, neural language systems exhibit variable accuracy in grasping temporal ordering entailed by prepositions like `before' and `after'. Similarly, in bAbI task 17, which concerns \textit{spatial reasoning}, \textsc{llm}-based systems fail to infer basic positional information that require interpreting the semantics of `to the left/right of', `above/below', etc. If such systems are inaccurate when dealing with common characteristics of the physical world, their performance doesn't improve when sentiments are considered: for instance, in SocialIQA \cite{sap-etal-2019-social}, given a context like `in the school play, Robin played a hero in the struggle to death with the angry villain', models are unable to consistently select `hopeful that Robin will succeed' over `sorry for the villain' when required to pick the correct answer to `how would others feel afterwards?'. It's not surprising that \textit{reasoning about emotional reactions} represents a difficult task for pure learning systems, when we consider that such form of inference is deeply rooted in the sphere of human experiences and social life, which involves a `layered' understanding of mental attitudes, intentions, motivations, emotions, and of the events that trigger them.\\
The qualitative analysis presented above suggests that neural models struggle to perform well in tasks that require high-level reasoning. But, are neuro-symbolic approaches sufficient to overcome these limitation? Latent expressions can augment training signals with sub-symbolic features derived from explicit semantic content, but knowledge infusion \textit{per se} doesn't determine how inference processes are conducted. Relevant work in this space shows how deep neural models can replicate logical reasoning \cite{ebrahimi2021capabilities, garcez2022neural}, but it doesn't follow that any form of logical reasoning that is provably reducible to learning algorithms, should also be systematically reduced to it -- this would be a requirement only for tightly-coupled neuro-symbolic systems \cite{kautz2022third, garcez2023neurosymbolic}. \\
Accordingly, in the next section we make the case for developing an AI framework where the \texttt{ACT-R} architecture is \textit{loosely-coupled} with neuro-symbolic components, to enable high-level reasoning.  

\section{Method}
\label{sec:method}

\begin{figure*}
   \centering
   \captionsetup{justification=centering}
    \includegraphics[scale=0.42]{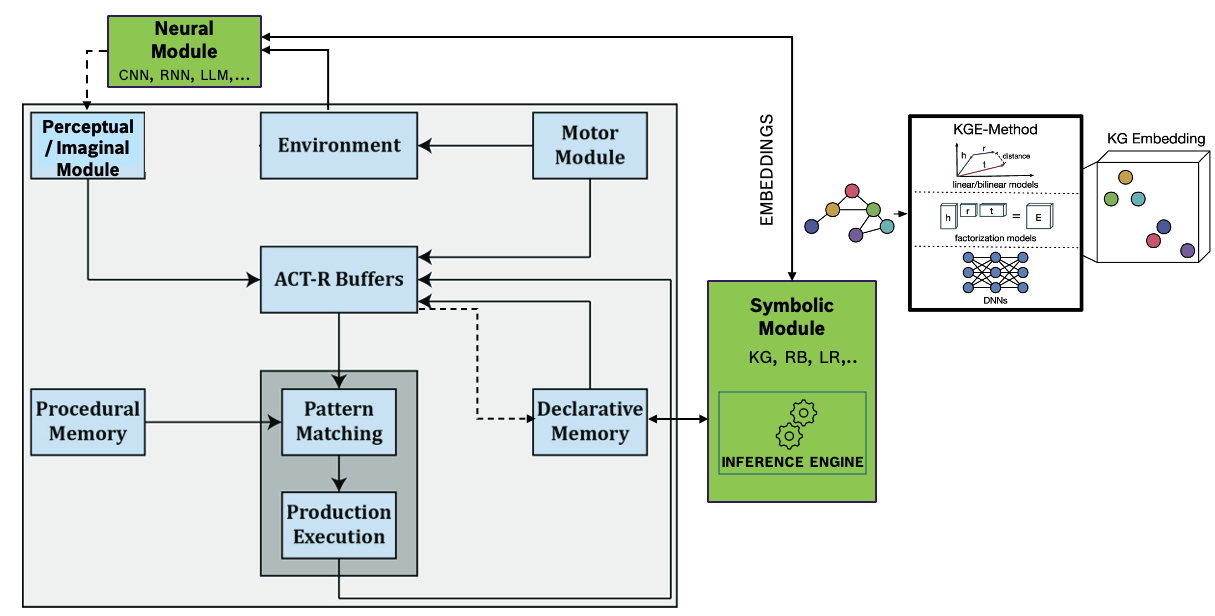}
    \caption{\textsc{act-r} integrated with neuro-symbolic modules.}
    \label{fig:ns-act-r}
\end{figure*}

Cognitive architectures attempt to capture at the computational level the invariant mechanisms of human cognition, including those underlying the functions of control, learning, memory, adaptivity, perception and action. \textsc{act-r} \cite{anderson1996act}, in particular, is designed as a hybrid modular framework including perceptual, motor and memory components, synchronized by a procedural module through limited capacity buffers. Over the years, \textsc{act-r} has accounted for a broad range of tasks at a high level of fidelity, reproducing aspects of complex human behavior, from everyday activities like event planning \cite{somers2020cognitive} and car driving \cite{cina2023categorized}, to highly technical tasks such as piloting an airplane \cite{chen2021developing}, and monitoring a network to prevent cyber-attacks \cite{ben2015ontology}. \textsc{act-r} has been used as a component in pipelines that include learning algorithms (e.g., biologically-inspired neural networks \cite{jilk2008sal}) and external semantic resources (e.g., \cite{oltramari2012using,emond2006wn}): along this line of research, we claim that integrating \textsc{act-r} -- or any compatible cognitive architecture -- with neuro-symbolic components is instrumental to enable high-level machine reasoning.\\ 
Figure \ref{fig:ns-act-r} provides a compact visualization of our proposed framework: the boxes in blue, enclosed in the grey rectangle, represent the default components of \texttt{ACT-R}, those in green the neuro-symbolic extensions. \\
The integration would occur along three main directions: 

\begin{itemize}
    \item \textbf{knowledge} $\leftrightsquigarrow$ \textbf{memory}: the external symbolic module, which can include background/domain knowledge graphs (\textsc{kg}), lexical resources (\textsc{lr}), rule bases (\textsc{rb}), and a suitable inference engine, is linked to the declarative memory. This is a two-way integration: the symbolic module can be \textit{read} or \textit{written} by \texttt{ACT-R}, where the latter operation is triggered when populating or pruning world knowledge is needed as part of task execution. 
    
    \item \textbf{neural $\rightsquigarrow$  perception}: the neural module, which can include convolutional, recurrent, long-short-term memory networks, generative models, etc., is trained, fine-tuned, or prompted with data processed from the environment, providing relevant patterns of information to the perceptual or imaginal module. This integration bypasses the direct connection holding -- in standard \texttt{ACT-R} -- between the perceptual module and the environment.\footnote{Such connection assumes symbolic representations of visual and auditory signals being available to the architecture through pre-processing.}
    
    \item \textbf{knowledge $\rightsquigarrow$ neural}: adequately-selected embedding mechanisms govern knowledge-infusion in the neural module, enabling knowledge-based contextualization of patterns of information distilled from the environment, which are subsequently channeled into \texttt{ACT-R} buffers. 
    
\end{itemize}

If the mutual connections between the two intertwined neuro-symbolic modules and \textsc{act-r} can be used to combine rich semantic contents with scalable learning functionalities, they don't \textit{per se} bring about high-level reasoning: this capability also requires two features of the integrated framework, namely the cognitive architecture's own procedural module and a proper inference engine in the external symbolic module. \\
The procedural module matches the content of the other module buffers and coordinates their activity using production rules, which are `condition-action' pairs tied to the task at hand. Productions use an utility-based computation to select, from a set of task-specific plausible rules, the single rule that is executed at any point in time. For instance, when building a recommendation system to support a mechanic in troubleshooting a car engine, a relevant situation that needs to be covered is a vehicle that doesn't start but has power; in this example, a high-utility production rule should capture the following heuristic: \textit{if the engine holds compression well, and the fuel system is working correctly, then the spark plugs should be checked}. The variables in these rule conditions would need to be filled with actual empirical observations and measurements, as it is often the case when cognitive architectures are applied in real-world scenarios: in our example, such evidence could be actually gathered by a real technician using the recommendation system in a human-machine-teaming fashion, a type of approach that falls under the `cognitive model as oracle' paradigm \cite{lebIEEE}.\\
The inference engine in the symbolic module is used to derive knowledge from assertions in the semantic resource of reference, a well-known feature of symbolic AI systems. What is important to stress here, is that  -- in our proposal -- this form of logic-based reasoning would realize two functions: 1) provide a combination of asserted and inferred knowledge that \textsc{act-r} declarative memory can process and pass to the production system; 2) support knowledge-infusion into neural modules. The first functionality would help to decouple basic forms of reasoning, e.g. temporal and spatial\footnote{E.g., Region-Connection-Calculus \cite{cohn1997qualitative} for spatial reasoning, Allen's axioms for temporal reasoning \cite{allen1994actions}.} , from cognitive assessments performed by the production system on conditional actions. Such feature makes our proposed system efficient, as \textsc{act-r} productions are not well-suited for logical reasoning. The second functionality would allow pre-training, fine-tuning, or prompting a neural-model on both asserted and inferred knowledge: this can provide \textsc{act-r} perceptual model with more informative patterns than just those obtained by processing raw data. \\
It's worth making a final consideration here: the framework introduced in this section is complemental to the body of work that investigates how neuro-symbolic systems can be leveraged to realize human-like cognitive reasoning \cite{garcez2008neural}: in our proposal, \textsc{act-r} is interfaced with neuro-symbolic components, whereas -- in the approaches reviewed by Garcez et al. -- neuro-symbolic frameworks are used to solve cognitive tasks. The difference lies on whether cognitive processes are considered \textit{first class citizens} or not. 

\section{Discussion: \\The Role of Generative AI in Cognitive Neuro-Symbolic Reasoning}
\label{sec:ca-gm}

As seen in the previous section, our proposed framework doesn't require or commit on a specific neural architecture. However, generative AI, and specifically large language models, will play an increasingly relevant role in enabling high-level reasoning based on \textit{cognitive neuro-symbolic systems}. In the next two sections we will briefly outline present research in this field, and sketch what we think are promising developments. 

\subsection{Related Work}
\label{sec:rw}
The importance of integrating cognitive mechanisms into data-driven AI systems has been recently acknowledged by one of the key figures in deep learning, Yann LeCun: in a position paper published in 2022 \cite{lecun2022path}, he described a biologically-inspired cognitive architecture, where a so-called \textit{configurator} orchestrates information provided by different modules, such as the \textit{perception module} and the \textit{world model module}, which replicate the functions emerging from prefrontal-cortical processes. Furthermore, a \textit{motivation model} -- designed to mimic the role of the amygdala in producing basic emotional states like pain and pleasure -- is used to compute intrinsic costs associated with current and future actions, a mechanism that is instrumental to inform predictive capabilities. It's relevant to point out that there has been extensive research on mapping cognitive architectures to brain areas/processes -- e.g., \cite{borst2015using} -- and that an established scientific community has been working on biologically-inspired approaches to cognitive architectures since the early 2000's (the BICA international conference has reached its \nth{14} edition\footnote{See: \url{https://bica2023.org/cfp/}}).\\
In line with the current trend of investigating computational models of cognition in the context of large-scale neural networks, a recent blog \cite{weng2023prompt} provides an overview of how \textsc{llm}s could be used to control autonomous agents. It goes beyond the scope of our contribution to review in detail the papers mentioned in the blog, but it's beneficial to highlight some of the most interesting topics. \\ 
In \cite{wei2022chain} the authors leverage \textit{chain-of-thought} prompting with \textsc{palm 540b} \cite{chowdhery2022palm} for task-decomposition: despite of their reported success, using prompting to generate fine-grained reasoning steps does not always yield consistent results, as shown by \cite{chen2023chatgpt} for different versions of \textsc{gpt-4} \cite{openai2023gpt}. The same work also indicates that, even when reasoning steps are correctly reproduced, they don't always match with the model selecting the correct solution/answer to a problem/question. Another paper surveyed in the blog \cite{park2023generative} focuses on using \textsc{gpt-4} to build \textit{believable agents} for a sandbox environment \footnote{Inspired by the video-game `The Sims': \url{https://www.ea.com/games/the-sims}}. According to the authors, cognitive architectures would not have the same degree of flexibility (and scalability) that modern generative models provide when building AI agents, as the former depend on hand-crafting rules, thus applicable only to narrow, closed-world contexts. However, this is a partial account of the state of the art: for instance, production compilation, \textsc{act-r}’s rule learning mechanism, allows to learn new, task-specific production rules that directly implement the relevant action(s) for a particular state \cite{taatgen2006cognitive}; moreover, to assess which stimuli from an environment are relevant for an agent to act upon, researchers have developed mechanisms like instance-based learning, a type of reinforcement learning \cite{sutton2018reinforcement}, which can be plugged into \textsc{act-r} \cite{gonzalez2003instance}. One may also question the claim on generative models' flexibility: in fact, the scope of such capability is not the real world, with its ever-changing situations, but rather some emerging patterns in the text-based training data, which are biased interpretations of the real world. Incidentally, this lack of `grounding' is also at the origin of \textsc{llm}s' widely-documented hallucination problem -- for an introduction to this phenomenon, see \cite{ji2023survey}.\footnote{There is an interesting analogy between Plato's Cave myth \cite{jowett1873dialogues}, where shadows projected on a blank wall were all that prisoners could use to understand reality, thus misinterpreting it, and \textsc{llm}s's generating inaccurate statements about the world, based on biased data patterns.}

    \begin{figure*}
   \centering
   \captionsetup{justification=centering}
    \includegraphics[scale=0.8]{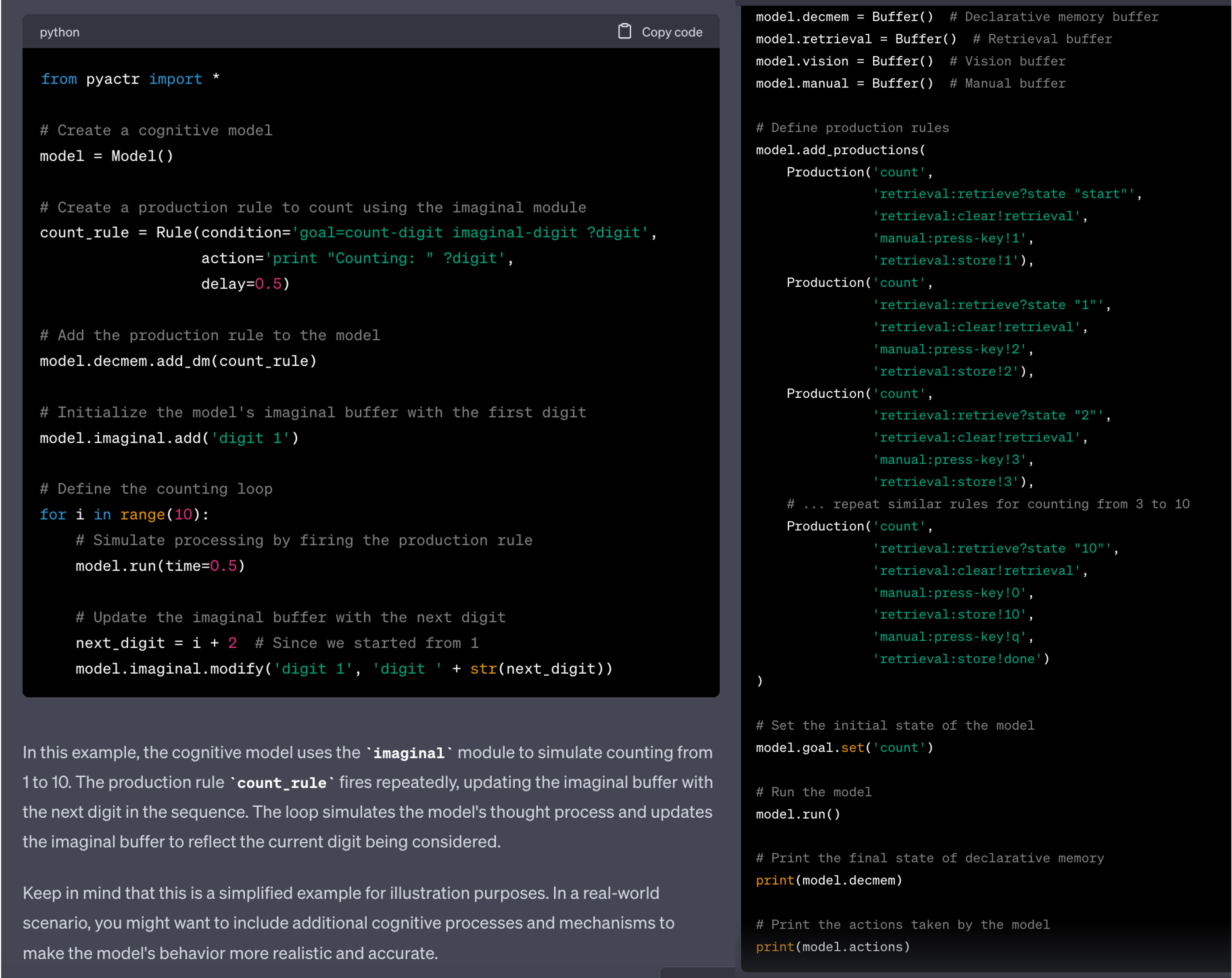}
    \caption{Without adequate instructions, \textsc{chat-gpt} opts for a compact model of counting from 1 to 10, which is not substantially distinct from a simple \texttt{for} loop (left-side). Interestingly, the chatbot suggests to \textit{include additional cognitive processes and mechanisms to make the model's behavior more realistic and accurate} (bottom-left): this is what actually happens when the \textsc{llm} is instructed to use all \textsc{act-r} modules (right-side). Far from being exhaustive, this example provides some evidence of the feasibility of scaling cognitive models via \textsc{llm}s.}
    \label{fig:counting}
\end{figure*}

\subsection{Future Work}
\label{sec:future}
As the overview in the previous section suggests, there are intrinsic limitations in utilizing a \textsc{llm} as \textit{orchestrator} for intelligent agents. In this regard, we can distill two main reasons for selecting a cognitive architecture over a \textsc{llm}: a) the inner functioning of the former is transparent, whereas the latter is a `black-box' \cite{castelvecchi2016can}; b) the former is  designed to replicate the invariant mechanisms of human cognition, the latter is engineered to produce human-grade linguistic \textit{behavior}, which cognitive properties can only be ascribed to. By and large, what the state-of-the-art suggests is that a synergistic integration of these cognitive architectures and \textsc{llm}s can help to maximize their relative strengths and mitigate their weaknesses, fostering the creation of more advanced AI systems, capable of high-level reasoning. In particular, (1) \textit{scaling cognitive models via} \textsc{llm}s and (2) \textit{prompt-engineering \textsc{llm}s with cognitive models} can be seen as novel approaches in this direction; they would actually be complemental, as (1) is a method to automatize the creation of cognitive models using generative AI, whereas (2) is a method to ground generative AI on computational artifacts that reflect principled cognitive theories. 

\begin{enumerate}
    \item \textbf{\textit{Scaling cognitive models via} \textsc{llm}s}. A cognitive architecture is a generic framework to develop cognitive models, which are, conversely, tied to specific tasks and domains: the process of developing cognitive models is still largely manual, and thus affected by lack of scalability. Because \textsc{llm}s have proven to be effective in generating code across a variety of programming languages \cite{gozalo2023chatgpt}, they could also be leveraged to produce software implementations of cognitive models. Initial experiments performed by asking \textsc{chat-gpt} to generate basic cognitive models using a novel library, i.e., \texttt{PyACT-R}\footnote{\url{https://github.com/jakdot/pyactr}}, show that the OpenAI's signature \textsc{llm} learns to correctly generate compact Python snippets, although it only makes marginal use of \textsc{act-r} modules and buffers. In order to achieve such level of sophistication in cognitive model design, \textsc{chat-gpt} needs to be prompted with relevant instructions about which mechanisms of a cognitive architecture it should use (see figure \ref{fig:counting}).

    \item \textbf{\textit{Prompt-engineering \textsc{llm}s} with cognitive models}. Using \textsc{llm}s in domain-specific applications requires either fine-tuning on a target dataset, or prompt-engineering with adequate contextual knowledge. In many use cases, well-curated data are either unavailable or too time-consuming to collect at scale, making the latter more convenient and efficient. When the goal is to turn a \textsc{llm} into a reliable decision support system, the `grounding' problem mentioned earlier also extends to the cognitive dimension: that is, such system would need to be based on shared interpretations of reality as well as on sound reasoning steps, from a cognitive-decisional standpoint. In fact, it'd be difficult to conceive such a system as trustworthy if hallucinations on both factual knowledge and on inferential mechanisms were widespread. To this end, prompting a \textsc{llm} with key steps of a cognitive model's reasoning process, the so-called \textit{trace}, would be instrumental to mitigate the second type of hallucinations. Such steps \textit{de facto} represent the introspective stages of a cognitive model, and of a cognitive neuro-symbolic reasoning system based on it. 

\end{enumerate}

\section{Conclusion}

In the current debate on the limits of deep neural networks, the split is oftentimes between those who think that \textit{more data} is the panacea, and those who support designing systems that integrate learning approaches with other processing elements, such as knowledge representation and reasoning, statistical algorithms, human-in-the-loop methods. In this paper, which echoes the second category, we made the case for adopting a cognitive approach to perform that integration, inspired by the results that architectures like \textsc{act-r} have produced, over the last decades, in replicating complex human tasks at the machine level. We described the main components of a \textit{cognitive neuro-symbolic reasoning system}, outlined their respective functionalities, and discussed related and future work in the area of generative AI. \\
At the end, to paraphrase Yoshua Bengio \cite{bengio}, we don’t assume or prove that using cognitive architectures is the only possibility to equip machines with high-level, human-like reasoning: however, through a diversity of scientific explorations, we'll increase our chances to find the ingredients we are missing.

\bibliography{aaai23}

\end{document}